\pdfoutput=1

\documentclass[11pt]{article}

\usepackage[final]{acl}

\usepackage{times}
\usepackage{latexsym}

\usepackage[T1]{fontenc}

\usepackage[utf8]{inputenc}

\usepackage{microtype}

\usepackage{inconsolata}

\usepackage{graphicx}

\title{Tokens for Learning, Tokens for Unlearning: Mitigating Membership Inference Attacks in Large Language Models via Dual-Purpose Training}

\author{Toan Tran, Ruixuan Liu\thanks{Corresponding author: ruixuan.liu2@emory.edu}, Li Xiong \\
  Emory University \\
  Atlanta, GA, USA \\
  \texttt{\{vtran29;ruixuan.liu2;lxiong\}@emory.edu} \\
  \textcolor{red}{Appeared at ACL'25 (Findings)}
}

\usepackage{pgfcalendar}
\newcount\myjuliandate
\newcount\myjuliantoday
\newcommand{\DaysTo}[3]{%
\pgfcalendardatetojulian{\year-\month-\day}{\myjuliantoday}%
\pgfcalendardatetojulian{#1-#2-#3}{\myjuliandate}%
\advance\myjuliandate by-\myjuliantoday\relax
\the\myjuliandate
}
\usepackage{multirow}
\usepackage{booktabs,arydshln}
\usepackage{xspace}

\newcommand{\gpt}{\mbox{GPT-2}\xspace}
\newcommand{\pythia}{\mbox{Pythia}\xspace}
\newcommand{\llama}{\mbox{Llama-2}\xspace}
\newcommand{\methodname}{\mbox{DuoLearn}\xspace}

\usepackage{xfrac}
\usepackage{xcolor}
\usepackage{listings}

\definecolor{codegreen}{rgb}{0,0.6,0}
\definecolor{codegray}{rgb}{0.5,0.5,0.5}
\definecolor{codepurple}{rgb}{0.58,0,0.82}
\definecolor{backcolour}{rgb}{0.95,0.95,0.92}

\lstdefinestyle{mystyle}{
    backgroundcolor=\color{backcolour},   
    commentstyle=\color{codegreen},
    keywordstyle=\color{magenta},
    numberstyle=\tiny\color{codegray},
    stringstyle=\color{codepurple},
    basicstyle=\ttfamily\footnotesize,
    breakatwhitespace=false,         
    breaklines=true,                 
    captionpos=b,                    
    keepspaces=true,                 
    numbers=left,                    
    numbersep=5pt,                  
    showspaces=false,                
    showstringspaces=false,
    showtabs=false,                  
    tabsize=2
}

\usepackage{amsmath}
\DeclareMathOperator*{\argmax}{arg\,max}
\DeclareMathOperator*{\argmin}{arg\,min}

\usepackage{graphics}
\usepackage{tikz}
\usetikzlibrary{shapes, arrows, positioning}
\usepackage{float}

\begin{document}
\maketitle
\begin{abstract}

Large language models (LLMs) have become the backbone of modern natural language processing but pose privacy concerns about leaking sensitive training data. Membership inference attacks (MIAs), which aim to infer whether a sample is included in a model's training dataset, can serve as a foundation for broader privacy threats. Existing defenses designed for traditional classification models do not account for the sequential nature of text data. As a result, they either require significant computational resources or fail to effectively mitigate privacy risks in LLMs. In this work, we propose \methodname, a lightweight yet effective empirical privacy defense for protecting training data of language models by leveraging token-specific characteristics. By analyzing token dynamics during training, we propose a token selection strategy that categorizes tokens into hard tokens for learning and memorized tokens for unlearning. Subsequently, our training-phase defense optimizes a novel dual-purpose token-level loss to achieve a Pareto-optimal balance between utility and privacy. Extensive experiments demonstrate that our approach not only provides strong protection against MIAs but also improves language modeling performance by around 10\% across various LLM architectures and datasets compared to the baselines.\footnote{The implementation code for DuoLearn is available at \url{https://github.com/Emory-AIMS/duolearn}}

\end{abstract}

\section{Introduction}

Large language models (LLMs) have become the foundation of modern natural language processing with a wide range of applications in various domains~\cite{llmevalsurvey}. The rapidly increasing deployment of LLMs raises serious concerns about data privacy~\cite{YAO2024100211}. LLMs have been shown to memorize the training data which can be later extracted by adversaries~\cite{carlini2023quantifying}. 
Membership inference attacks (MIAs)~\cite{MIAShokri2017, li2024privacyllms} aim to infer whether a sample is included in a model's training data, serving as the foundation of broader privacy threats~\cite{carlini2021extracting}.

Due to the importance of understanding and mitigating MIAs, a significant amount of research has been conducted to design MIA defenses~\cite{miasurvey}. 
However, most defenses focus on general machine learning models for classification tasks and do not account for the sequential nature of text data, while advanced MIAs for LLMs have leveraged this property.
For example, the series of Min-K works~\cite{zhang2025mink, shi2024detecting} uses the token-level loss on outlier tokens and significantly enhance MIAs for LLMs.
Thus, conventional data sanitization or regularization techniques have limited defense effectiveness~\cite{kandpal2022deduplicating, precurious}.
Even though the classic differentially private (DP) training algorithm~\cite{abadi2016deep} provides a strong defense, this approach comes at the inevitable cost of increased computation and reduced utility~\cite{li2022does, bu2023groupwise}, which may not be desirable when the model trainer serves as the defender.


In this paper, we propose a defense mechanism for membership inference attacks on LLMs -- \methodname. 
A recent study~\cite{lin2024not} reveals that using a carefully selected subset of tokens during training can match or even surpass the performance of using all tokens in language modeling. In the meantime, MIAs mainly exploit loss-based signals associated with a sample~\cite{neighbourattack, Carlini2020ExtractingTD}. We observe that during training, some certain tokens carry stronger MIA signals than others, making the sample vulnerable to MIAs. Thus, we leverage the token sequence nature of LLMs and propose a dynamic token selection strategy during training to proactively identify and categorize tokens into hard tokens (those the model struggles with)  and memorized tokens (those with strong signals for MIA risks). Accordingly, we design a dual-objective loss function that performs learning via gradient descent on the hard tokens and unlearning via gradient ascent on the memorized tokens simultaneously in one backward pass, which makes the model learn useful information but not memorize specific training samples.
Our contributions can be summarized as follows:
\begin{itemize}
    \item We propose a dynamic token selection strategy that identifies hard tokens and memorized tokens during training, which provides insights for investigating language modeling and memorization.
    \item We propose a simple but effective dual-objective training to perform learning over hard tokens and unlearning over memorized tokens, for mitigating privacy risk while maintaining model utility with small computing cost.
    \item We empirically demonstrate the effectiveness of the proposed defense mechanism across various LLM architectures and datasets. Our results show that our defense mechanism can provide robust privacy protection against MIAs with minimal degradation on language modeling performance.
\end{itemize}

\section{Related Works}
\subsection{MIAs on LLMs}
Membership inference attacks are a crucial privacy threat to machine learning models. There are a significant number of MIAs proposed for traditional classification models~\cite{miasurvey}. \citet{MIAShokri2017} introduce MIAs via training a binary classification model over behaviors collected from shadow models.
\citet{8429311} connect MIAs to the overfitting phenomenon and propose to use cross entropy loss as an MIA signal.  However, due to the significant differences between LLMs and traditional classification models, some of these attacks are not applicable to LLMs, while others, though feasible, have limited attack performance. Therefore, there are non-trivial efforts to design suitable MIAs for LLMs. \citet{Carlini2020ExtractingTD} calibrate the sample loss with zlib entropy and reference models. \citet{neighbourattack} generate synthetic neighboring samples for each target sample then calibrate the loss of the target sample with the averaged loss of its neighboring samples as the MIA signal. \citet{shi2024detecting} consider only top $K$ lowest token losses for the MIA signal, while \citet{zhang2025mink} perform z-score normalization for token losses, using the token vocabulary's mean and standard deviation, then select top $K$ z-scores. \citet{fu2024membership} prompt the target LLM to generate a dataset which is used to train a reference attack model. \citet{duan2024membership} conduct systematic evaluations of MIAs on the pretrained LLMs. \citet{hayes2025strong} scale reference-based MIAs on large-scale pretraining settings. \citet{puerto2025smia} consider various scales of membership from sentences to collections of documents. \citet{precurious} design a privacy backdoor that can increase the membership inference risks. \citet{feng2025exposing} investigate MIAs on preference data used for post-training alignment.

\subsection{LLM Memorization}
The billion-parameter scale enhances LLM capabilities but also magnifies the privacy concerns. \citet{Carlini2020ExtractingTD, carlini2023quantifying} demonstrate that LLMs can memorize parts of their training data. There is a risk that LLMs may generate the training data when prompted appropriately. These are known as \textit{exact memorization} which can be utilized by the adversaries to extract the exact training data. \citet{nasr2025scalable} demonstrated that the LLM safety alignment fails to mitigate the privacy risks. It is feasible to undo the safety alignment via fine-tuning and the adversaries can prompt the LLM to generate its training data.

\subsection{Defenses Against MIAs}
Overfitting is the root of membership inference risks~\cite{MIAShokri2017}. 
While regularization methods such as weight decay and dropout~\cite{dropout} mitigates overfitting and slightly reduces the membership inference risks in the traditional classification models~\cite{272134}, they are not sufficient to prevent memorization in LLMs~\cite{tirumala2022memorization, lee2022deduplicating}. 
Other defenses which leverage adversarial training~\cite{Miladadvereg} or ensemble architecture of models~\cite{280000} are infeasible for LLMs due to the expensive computing cost. 

Generally, in the context of LLMs, there are still limited number of works on defense mechanisms against MIAs and memorization. There are two main approaches: sanitizing training data and differential privacy (DP). \citet{p2022textanonymbench} propose a practical method to protect Personally Identifiable Information (PII) by detecting and replacing PII with anonymized tokens. \citet{shi2022just} sanitize the PII tokens and  pretrain on the sanitized data before conducting DP based fine-tuning on the original data. \citet{10179300} demonstrate the effectiveness of sentence-level DP in mitigating the risks of leaking PII. These PII protection methods are effective but may not be sufficient to protect against MIAs because for each sample, the number of PII tokens is usually small~\cite{llmpbe}. \citet{liu2024exp} propose a method to perturb the training texts by leveraging memorization triggers that can effectively protect a small fraction of the training data against MIAs. Deduplicating the training corpus can reduce the risks of MIAs but not entirely eliminate them~\cite{kandpal2022deduplicating}.

The second popular approach conducts training/fine-tuning with Differentially-Private Stochastic Gradient Descent (DPSGD). \citet{li2022large,yu2022differentially,llmpbe,amit2024sok} show LLMs are strong differentially private learners. 
\citet{lowy2024dptheo} theoretically prove that DP with a loose privacy budget can defend against MIAs. 
Despite efforts to improve the computing efficiency of DPSGD~\cite{bu2023groupwise}, DP inherently introduces computational overhead, architectural constraints, and significant utility trade-off at scale. \citet{mckenna2025scalinglaws} explore the scaling laws of DP LLMs and reveal challenges especially about the optimal training data size. To avoid the computational overhead and utility tradeoff of using DP on LLMs, \citet{hans2024be} proposes a non-DP practical masking mechanism, called Goldfish, that performs pseudo-random token masking for loss calculation to prevent memorization.
Our method is also an empirical defense in the training stage.
\section{How Do Tokens Contribute to Membership Inference Risks?}
\label{sec:analysis}
Compared to conventional classification problems, membership inference attacks in language modeling have significant differences. In particular, each query in traditional classification models requires only one prediction. On the other hand, each query to language models involves multiple token predictions due to the sequential nature of text. 
This difference yields a question that how tokens contribute to overall sample-level membership inference risks. To answer this question, we perform a token-level analysis of membership inference risks. We calculate the MIA signal for each token as its prediction loss calibrated by a reference model~\cite{Carlini2020ExtractingTD}.
A smaller signal value indicates that the model has a significantly higher confidence than the reference model on predicting the token.

 \begin{figure}[htp]
    \centering
    \includegraphics[width=0.46\linewidth]{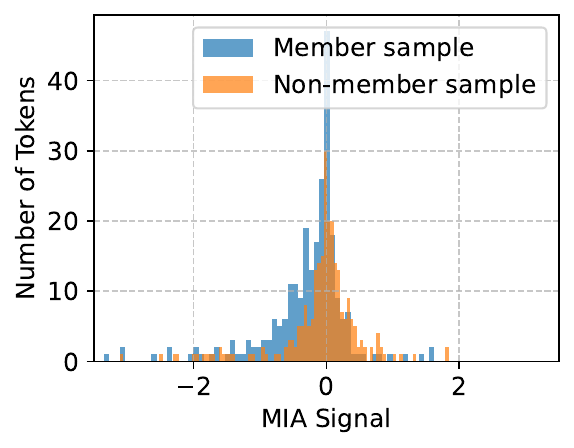}
    \includegraphics[width=0.512\linewidth]{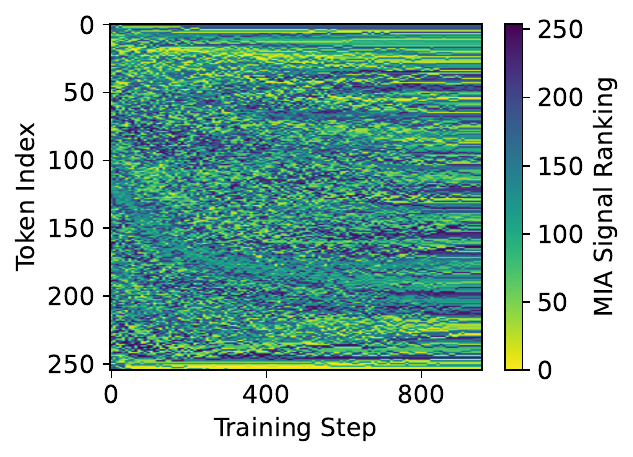}
    \caption{Token-level MIA signal analysis. The left figure presents the histogram of the MIA signals across tokens at the end of training, while the right figure illustrates the MIA signal ranking of tokens during training.}
    \label{fig:per-token-signal}
\end{figure}

Figure~\ref{fig:per-token-signal} (left) illustrates the histogram of MIA signal values for the tokens of a member sample and a non-member sample (see Figure~\ref{fig:add-per-token-loss} in Appendix~\ref{sec:app-analysis} for additional histograms). 
The non-member sample distribution centers around zero, while the member sample skews to the negative side. Consequently, the average aggregated MIA signal is below zero for the member but around zero for the non-member, leading to membership inference risks. Moreover, the MIA signal values vary for different tokens, so some tokens contribute more to the membership inference risks than the others. Figure~\ref{fig:per-token-signal} (right) illustrates the MIA signal ranking of tokens of a member sample over training steps (see Figure~\ref{fig:add-per-token-dynamics} in Appendix~\ref{sec:app-analysis} for additional samples). There is a complex changing dynamic in ranking between tokens before it becomes more stable at the end when the training converges. Overall, the analysis suggests that the sample-level membership inference risk in language modeling stem from the cumulative effect of many tokens. This poses challenges for defense methods, as they need token-level granularity to isolate and mitigate specific sources of leakage. Additionally, it is non-trivial to develop a defense method that widely affects a large number of tokens without disrupting the complex token dependencies essential for model utility.


\section{Proposed Methodology -- \methodname}
\label{sec:method}
\begin{figure*}[htp]
    \centering
    \includegraphics[width=0.9\textwidth]{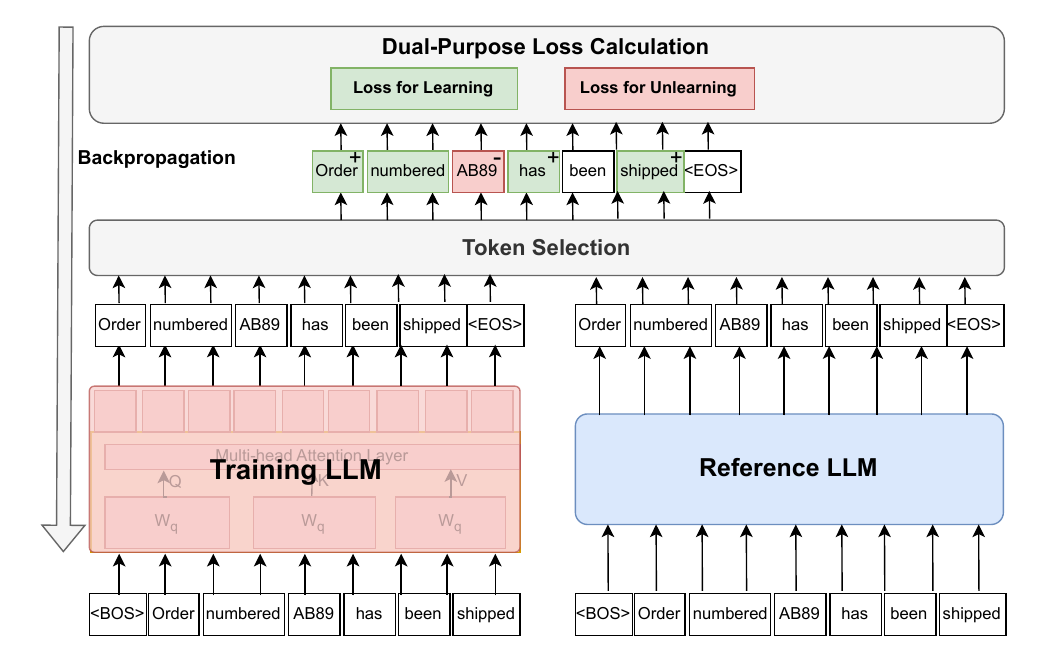}
    \caption{\methodname overview. First, the tokens are passed through the training LLM and reference LLM. They are then categorized into hard tokens (in green) and memorized tokens (in red). At the end, a dual-purpose loss is applied which achieves two targets: learning on the hard tokens while unlearning for the memorized tokens.}
    \label{fig:duolearn}
\end{figure*}

Motivated by the analysis, we propose \methodname -- a training framework that dynamically identifies hard tokens (those with higher calibrated losses) for learning and memorized tokens (those with lower calibrated loss or stronger MIA signals) for unlearning simultaneously. This way, the model learns useful information without memorizing specific training samples.

\noindent 
\textbf{Overview}. 
We assume the model trainer is the defender and the goal is to mitigate the privacy risk of the training data in the trained model.
We further assume the trainer can get access to an auxiliary dataset for better calibrating the MIA signals, which can be a disjoint subset drawn from the same distribution of the training data. The general training process is illustrated in Figure~\ref{fig:duolearn}.
First, we train a reference model with the auxiliary dataset, which is feasible for the trainer based on previous works~\cite{lin2024not, 2022PrioritizedTraining, xie2023doremi}.
Then, during training of the target model, we use the token losses of the current training model calibrated by the reference model to dynamically identify hard tokens and memorized tokens in each training iteration. 
A dual-purpose loss function is used to keep the model simultaneously learning on hard and necessary tokens to enhance model utiilty while unlearning on memorized tokens to mitigate MIA risks.

\noindent \textbf{Reference Modeling}. Reference model ($\theta_\text{ref}$) shares an identical architecture with the training model ($\theta$). We fine-tune a reference model on a small portion of the original dataset (denoted as $\mathcal{T}_\text{aux}$) that can reflect the desired data distribution by standard causal language modeling (CLM), i.e., implementing next-token-prediction cross entropy loss ($\mathcal{L}_{CE}$):
\begin{equation*}
  \resizebox{\linewidth}{!}{$\mathcal{L}_{CE}(\theta_\text{ref}; \mathcal{T}_\text{aux}) = - \frac{1}{|\mathcal{T}_\text{aux}|} \sum_{t_i \in \mathcal{T}_\text{aux}} \log P(t_i| t_{<i}; \theta_\text{ref})$}.
\end{equation*}
\noindent \textbf{Token Selection}. 
As our previous analysis, tokens contribute differently in membership inference risks.
Thus, considering all tokens equally is not optimal for privacy defense with respect to the utility and privacy trade-off.
\methodname defines two sets of tokens: hard tokens ($\mathcal{T}_{h}$) and memorized tokens ($\mathcal{T}_m$). Hard tokens are the tokens that the current training models ($\theta$) have difficulty predicting, while memorized tokens are the tokens that the model has already memorized. To identify these two sets of tokens, we propose a token selection mechanism based on the prediction loss of each token calibrated by the reference model. We implement the score $s(t_i)$ for each token $t_i$ which is the difference between the cross-entropy loss of the training model and the reference model, as used in previous works~\cite{lin2024not, 2022PrioritizedTraining}:
\[ s(t_i) = \log P(t_i| t_{<i}; \theta_\text{ref}) - \log P(t_i| t_{<i}; \theta). \]

The tokens with the highest scores are considered hard tokens $\mathcal{T}_{h}$ (highest calibrated loss), while the tokens with the lowest scores are considered memorized tokens $\mathcal{T}_m$ (lowest calibrated loss and strongest MIA signals). Let $\mathcal{T}$ be the set of all tokens in a batch. We select top $K_h$ hard tokens and bottom $K_m$ memorized tokens to form $\mathcal{T}_{h}$ and $\mathcal{T}_m$, respectively. Additionally, we introduce a threshold $\tau$ to filter out neutral tokens from $\mathcal{T}_m$ which have scores close to zero or greater than zero, as these are not considered memorized. The token selection process is formulated as follows:
\[ \mathcal{T}_{h} = \argmax_{S, |S|=K_h}\{{s(t_i) | t_i \in \mathcal{T}}\}  \]
\[ \mathcal{T}_{m} = \argmin_{S, |S| \leq K_m}\{{s(t_i) | t_i \in \mathcal{T}, s(t_i) \leq \tau}\} \]

\noindent \textbf{Dual-Purpose Loss}. We introduce a dual-purpose loss function designed to improve model performance on hard tokens while mitigating overfitting on memorized tokens. This loss function combines two components: the learning loss and the unlearning loss. The learning loss is the standard causal language modeling (CLM) loss applied to the hard tokens $\mathcal{T}_{h}$. The unlearning loss, in contrast, is the negative CLM loss applied to the memorized tokens $\mathcal{T}_{m}$, effectively performing gradient ascent. The dual-purpose loss is defined as follows, where $\alpha > 0$ is a hyper-parameter that balances the learning and unlearning losses
\[ \mathcal{L}_{dual}(\theta) = \mathcal{L}_{CE}(\theta; \mathcal{T}_{h}) - \alpha \cdot \mathcal{L}_{CE}(\theta; \mathcal{T}_{m}). \]


    
\section{Experiments and Results}
\subsection{Experiment Settings}

\begin{table*}[h]
  \centering
  \resizebox{0.9\textwidth}{!}{\begin{tabular}{cl|ccccc|ccccc}
  \toprule[1pt]
   \multirow{3}{*}{\textbf{LLM}}  & \multirow{3}{*}{\textbf{Method}} &  \multicolumn{5}{c|}{\textbf{Wikipedia}} & \multicolumn{5}{c}{\textbf{CC-news}} \\ \cmidrule(lr){3-7}  \cmidrule(lr){8-12}
    &  & PPL & Loss & Ref & Min-k & \multicolumn{1}{c|}{Zlib} & PPL & Loss & Ref & Min-k & Zlib \\ \midrule
    \multirow{4}{*}{GPT2} & \textit{Base} & \textit{34.429} & \textit{0.473} & \textit{0.513} & \textit{0.446} & \textit{0.497} & \textit{29.442} & \textit{0.505} & \textit{0.498} & \textit{0.520} & \textit{0.500} \\ 
    \multirow{4}{*}{124M} & FT & \textbf{12.729} & 0.577 & 0.967 & 0.489 & 0.544 & \textbf{21.861} & 0.607 & 0.855 & 0.549 & 0.569 \\
    & Goldfish & 12.853 & 0.565 & 0.954 & 0.486 & 0.537 & 21.902 & 0.608 & 0.855 & 0.547 & 0.570 \\
    & DPSGD & 18.523 & 0.463 & 0.536 & \textbf{0.448} & 0.491 & 26.022 & 0.507 & 0.513 & \textbf{0.521} & 0.502 \\
    & \methodname & 13.628 & \textbf{0.454} & \textbf{0.463} & 0.470 & \textbf{0.485} & 23.733 & \textbf{0.502} & \textbf{0.495} & 0.529 & \textbf{0.499} \\ \midrule
    
    \multirow{4}{*}{Pythia} & \textit{Base} & \textit{10.287} & \textit{0.466} & \textit{0.503} & \textit{0.464} & \textit{0.489} & \textit{13.973} & \textit{0.507} & \textit{0.512} & \textit{0.528} & \textit{0.501}\\ 
    \multirow{4}{*}{1.4B} & FT & \textbf{6.439} & 0.578 & 0.985 & 0.484 & 0.557 & 11.922 & 0.602 & 0.857 & 0.541 & 0.574 \\
    & Goldfish & 6.465 & 0.564 & 0.981 & 0.482 & 0.546 & \textbf{11.903} & 0.609 & 0.862 & 0.543 & 0.579 \\
    & DPSGD & 7.751 & 0.469 & 0.524 & \textbf{0.462} & 0.488 & 13.286 & 0.512 & 0.531 & 0.528 & 0.503 \\
    & \methodname & 6.553 & \textbf{0.468} & \textbf{0.485} & 0.472 & \textbf{0.485} & 12.670 & \textbf{0.501} & \textbf{0.460} & \textbf{0.524} & \textbf{0.499} \\ \midrule
    
    \multirow{4}{*}{Llama-2} & \textit{Base} & \textit{7.014} & \textit{0.458} & \textit{0.491} & \textit{0.476} & \textit{0.488} & \textit{9.364} & \textit{0.505} & \textit{0.495} & \textit{0.516} & \textit{0.503} \\ 
    \multirow{4}{*}{7B} & FT & \textbf{3.830} & 0.524 & 0.936 & 0.494 & 0.530 & \textbf{6.261} & 0.559 & 0.798 & 0.536 & 0.548 \\
    & Goldfish & 3.839 & 0.518 & 0.929 & 0.492 & 0.525 & 6.280 & 0.552 & 0.780 & 0.533 & 0.541 \\
    & DPSGD & 4.490 & 0.466 & 0.516 & \textbf{0.470} & 0.487 & 6.777 & 0.509 & 0.538 & 0.523 & 0.504 \\
    & \methodname & 4.006 & \textbf{0.458} & \textbf{0.440} & 0.473 & \textbf{0.480} & 6.395 & \textbf{0.507} & \textbf{0.482} & \textbf{0.518} & \textbf{0.500} \\
    \bottomrule[1pt]
  \end{tabular}}
  \caption{Overall Evaluation: Perplexity (PPL) and AUC scores of the MIAs with different signals (Loss/Ref/Min-k/Zlib). For all metrics, the lower the value, the better the result. \textit{Base} in the method column indicates the pretrained LLMs without fine-tuning, thus it indicates lower bound for both utility and privacy risk.}
  \label{tab:main_result}
\end{table*}

\textbf{Datasets}. We conduct experiments on two datasets: CC-news\footnote{\href{https://huggingface.co/datasets/vblagoje/cc_news}{Huggingface: vblagoje/cc\_news}} and Wikipedia\footnote{\href{https://huggingface.co/datasets/legacy-datasets/wikipedia}{Huggingface: legacy-datasets/Wikipedia}}. CC-news is a large collection of news articles which includes diverse topics and reflects real-world temporal events. Meanwhile, Wikipedia covers general knowledge across a wide range of disciplines, such as history, science, arts, and popular culture.\\
\textbf{LLMs}: We experiment on three models including \gpt~(124M)~\cite{gpt2radford}, \pythia~(1.4B)~\cite{pythia}, and \llama~(7B)~\cite{llama2touvron2023}. This selection of models ensures a wide range of model sizes from small to large that allows us to analyze scaling effects and generalizability across different capacities. \\
\textbf{Evaluation Metrics}. For evaluating language modeling performance, we measure perplexity (PPL), as it reflects the overall effectiveness of the model and is often correlated with improvements in other downstream tasks~\cite{kaplan2020scalinglaws, lmsfewshot}. For defense effectiveness, we consider the attack area under the curve (AUC) value and True Positive Rate (TPR) at low False Positive Rate (FPR). In total, we perform 4 MIAs with different MIA signals. Given the sample $x$, the MIA signal function $f$ is formulated as follows: \\
$\bullet$ Loss~\cite{8429311} utilizes the negative cross entropy loss as the MIA signal. 
    \[f_\text{Loss}(x) = -\mathcal{L}_\text{CE}(\theta; x) \]
$\bullet$ Ref-Loss~\cite{Carlini2020ExtractingTD} considers the loss differences between the target model and the attack reference model. To enhance the generality, our experiments ensure there is no data contamination between the training data of the target, reference, and attack models.
    \[f_\text{Ref}(x) = \mathcal{L}_\text{CE}(\theta_\text{ref}; x) - \mathcal{L}_\text{CE}(\theta; x) \]
$\bullet$ Min-K~\cite{shi2024detecting} leverages top K tokens that have the lowest probabilities.
    \[f_\text{min-K}(x) = \frac{1}{|\text{min-K(x)}|} \sum_{t_i \in \text{min-K(x)}} \log P(t_i|t_{<i};\theta) \]
$\bullet$ Zlib~\cite{Carlini2020ExtractingTD} calibrates the loss signal with the zlib compression size.
    \[ f_\text{zlib}(x) = -\mathcal{L}_\text{CE}(\theta; x) / \text{zlib}(x) \]

\noindent \textbf{Baselines}. We present the results of four baselines. \textit{Base} refers to the pretrained LLM without fine-tuning. \textit{FT} represents the standard causal language modeling without protection. \textit{Goldfish}~\cite{hans2024be} implements a masking mechanism. \textit{DPSGD}~\cite{abadi2016deep, yu2022differentially} applies gradient clipping and injects noise to achieve  sample-level differential privacy.

\noindent \textbf{Implementation}. We conduct full fine-tuning for \gpt and \pythia. For computing efficiency, \llama fine-tuning is implemented using Low-Rank Adaptation (LoRA)~\cite{hu2022lora} which leads to \textasciitilde4.2M trainable parameters. Additionally, we use subsets of 3K samples to fine-tune the LLMs. The data used to train \methodname's reference model is disjoint from either the target model's or the reference attack model's training data. We present additional implementation details in Appendix~\ref{sec:app-implementation}.

\subsection{Overall Evaluation}
Table~\ref{tab:main_result} provides the overall evaluation compared to several baselines across large language model architectures and datasets. Among these two datasets, CCNews is more challenging, which  leads to higher perplexity  for all LLMs and fine-tuning methods. Additionally, the reference-model-based attack performs the best and demonstrates high privacy risks with attack AUC on the conventional fine-tuned models at 0.95 and 0.85 for Wikipedia and CCNews, respectively. Goldfish achieves similar PPL to the conventional FT method but fails to defend against MIAs. This aligns with the reported results by \citet{hans2024be} that Goldfish resists exact match attacks but only marginally affects MIAs. DPSGD provides a very strong protection in all settings (AUC < 0.55) but with a significant PPL tradeoff. Our proposed \methodname guarantees a robust protection, even slightly better than DPSGD, but with a notably smaller tradeoff on language modeling performance. For example, on the Wikipedia dataset, \methodname delivers perplexity reduction by 15\% to 27\%. Moreover, Table~\ref{tab:tpr} (Appendix~\ref{sec:app-add-res}) provides the TPR at 1\% FPR. Both DPSGD and \methodname successfully reduce the TPR to $\sim$0.02 for all LLMs and datasets. \textit{Overall, across multiple LLM architectures and datasets, \methodname consistently offers ideal privacy protection with  little trade-off in language modeling performance.}

\begin{figure}[h]
    \centering
    \includegraphics[width=\linewidth]{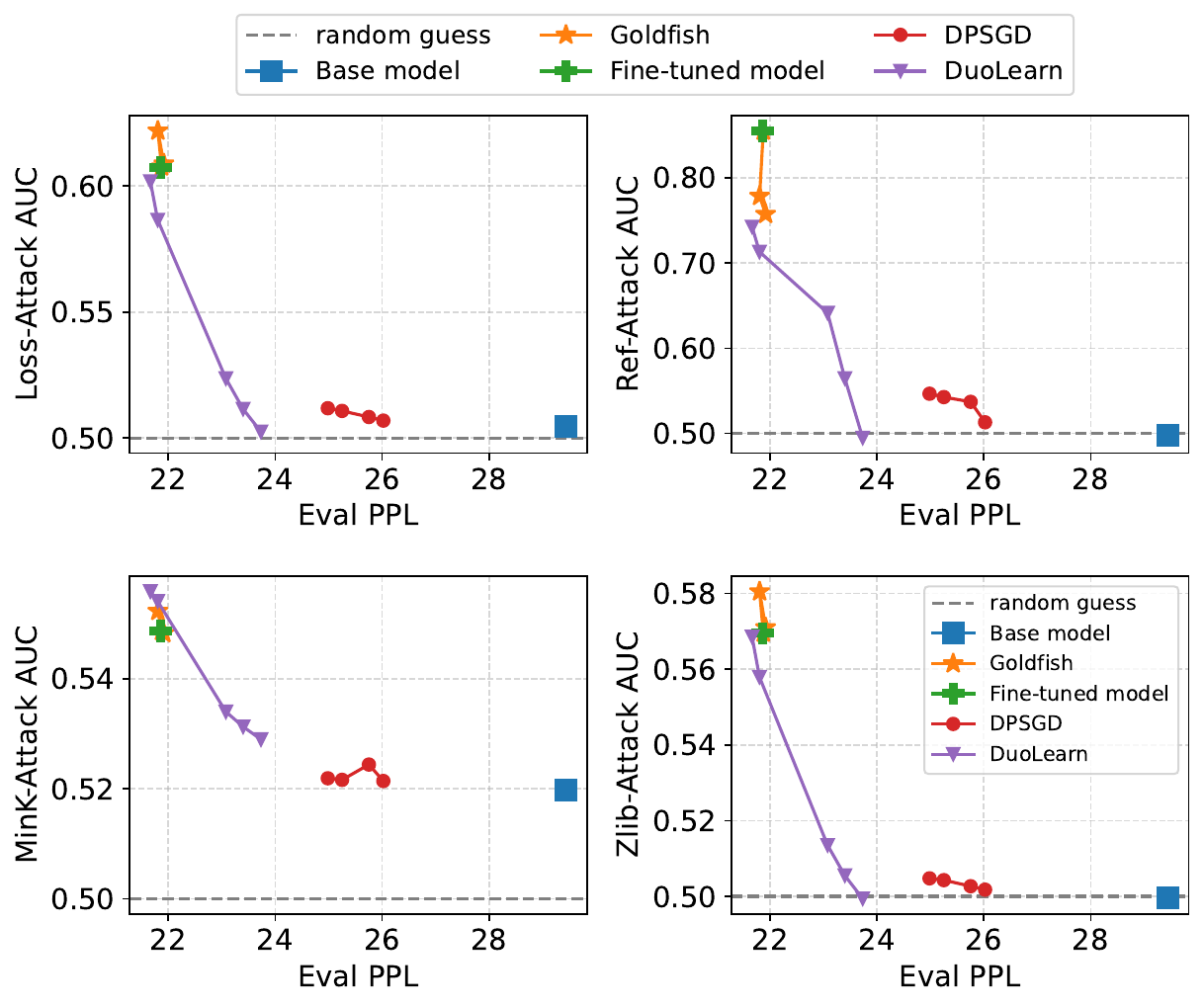}
    \caption{Privacy-utility trade-off of the methods while varying hyper-parameters. The Goldfish masking rate is set to 25\%, 33\%, and 50\%. The privacy budget $\epsilon$ of DPSGD is evaluated at 8, 16, 50, and 100. The weight $\alpha$ of \methodname is varied at 0.0, 0.1, 0.2, 0.5, and 0.8.}
    \label{fig:priv-ult-tradeoff}
\end{figure}

\noindent \textbf{Privacy-Utility Trade-off.}
To investigate the privacy-utility trade-off of the methods, we vary the hyper-parameters of the fine-tuning methods. Particularly, for DPSGD, we adjust the privacy budget $\epsilon$ from (8, 1e-5)-DP to (100, 1e-5)-DP. We modify the masking percentage of Goldfish from 25\% to 50\%. Additionally, we vary the loss weight $\alpha$ from 0.2 to 0.8 for \methodname. Figure~\ref{fig:priv-ult-tradeoff} depicts the privacy-utility trade-off for GPT2 on the CCNews dataset. Goldfish, with very large masking rate (50\%), can slightly reduce the risk of the reference attack but can increase the risks of other attacks. By varying the weight $\alpha$, \methodname offers an adjustable trade-off between privacy protection and language modeling performance. \methodname largely dominates DPSGD and improves the language modeling performance by around 10\% with the ideal privacy protection against MIAs.

\subsection{Ablation Study}
\begin{figure}[htp]
    \centering
    \includegraphics[width=0.35\textwidth]{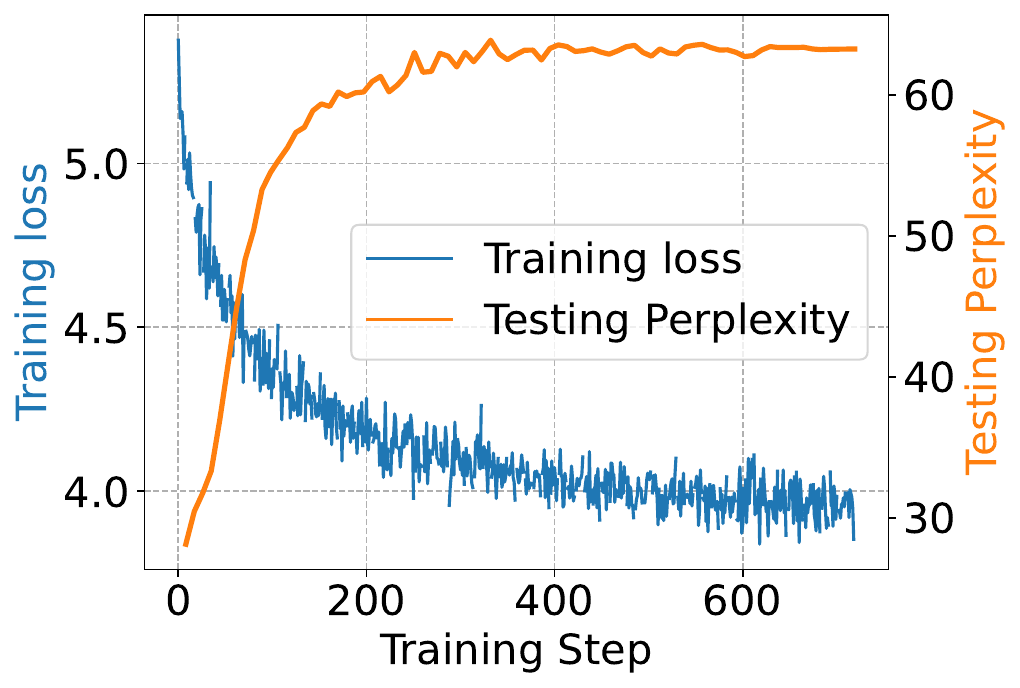}
    \caption{Training Loss and Test Perplexity of \methodname without a reference model.
    }
    \label{fig:ppl-auc-noref}
\end{figure}

\textbf{\methodname without reference models.} To study the impact of the reference model, we adapt \methodname to a non-reference version which directly uses the loss of the current training model (i.e., $s(t_i) = \mathcal{L}_{CE}(\theta; t_i)$) to select the learning and unlearning tokens. This means the unlearning tokens are the tokens that have smallest loss values. Figure~\ref{fig:ppl-auc-noref} presents the training loss and testing perplexity. There is an inconsistent trend of the training loss and testing perplexity. Although the training loss decreases overtime, the test perplexity increases. This result indicates that identifying appropriate unlearning tokens  without a reference model is challenging and conducting unlearning on an incorrect set hurts the language modeling performance.

\begin{figure}[htp]
    \centering
    \includegraphics[width=0.3\textwidth]{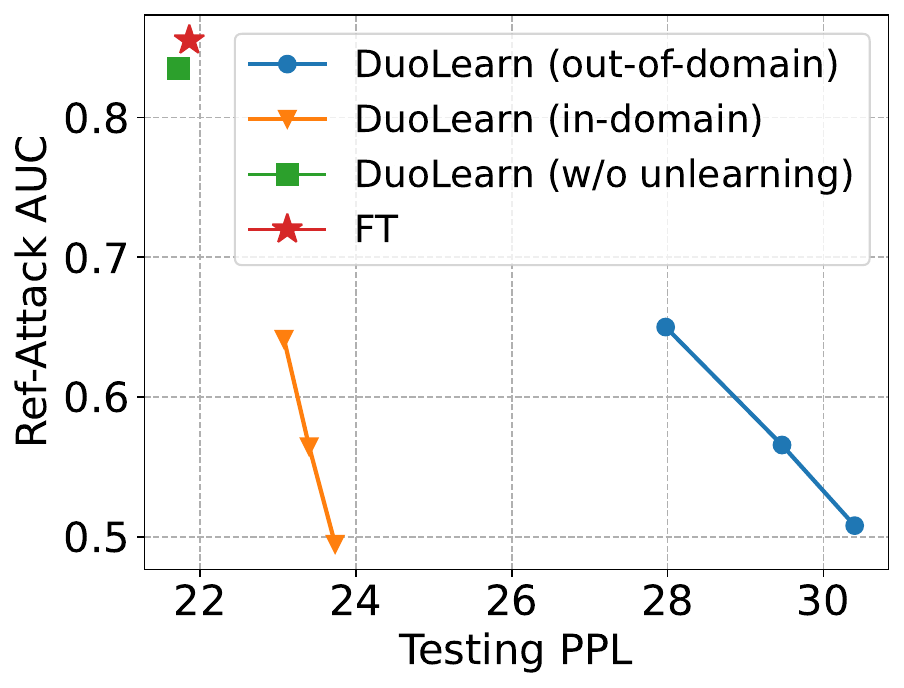}
    \caption{Privacy-utility trade-off of \methodname with different settings: in-domain reference model, out-domain reference model, and without unlearning}
    \label{fig:ppl-auc-base-woasc}
\end{figure}

\noindent \textbf{\methodname with out-of-domain reference models.} To examine the influence of the distribution gap in the reference model, we replace the in-domain trained reference model with the original pretrained base model. 
Figure~\ref{fig:ppl-auc-base-woasc} depicts the language modeling performance and privacy risks in this study. \methodname with an out-of-domain reference model can reduce the privacy risks but yield a significant gap in language modeling performance compared to \methodname using an in-domain reference model.

\noindent \textbf{\methodname without Unlearning.} To study the effects of unlearning tokens, we implement \methodname which use the first term of the loss only ({$\mathcal{L}_{\theta} = \mathcal{L}_{CE}(\theta; \mathcal{T}_h)$}). Figure~\ref{fig:ppl-auc-base-woasc} provides the perplexity and MIA AUC scores in this setting. Generally, without gradient ascent, \methodname can marginally reduce membership inference risks while slightly improving the language modeling performance. The token selection serves as a regularizer that helps to improve the language modeling performance. Additionally, tokens that are learned well in previous epochs may not be selected in the next epochs. This slightly helps to not amplify the memorization on these tokens over epochs.

\subsection{Training Dynamics}
\begin{figure*}[htp]
    \centering
    \includegraphics[width=0.305\linewidth]{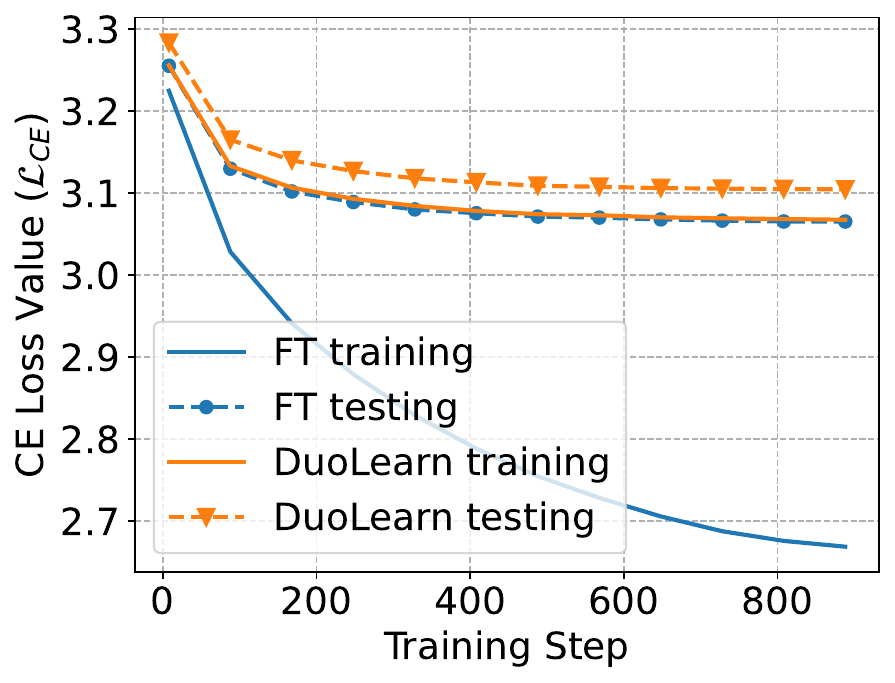}
    \includegraphics[width=0.305\linewidth]{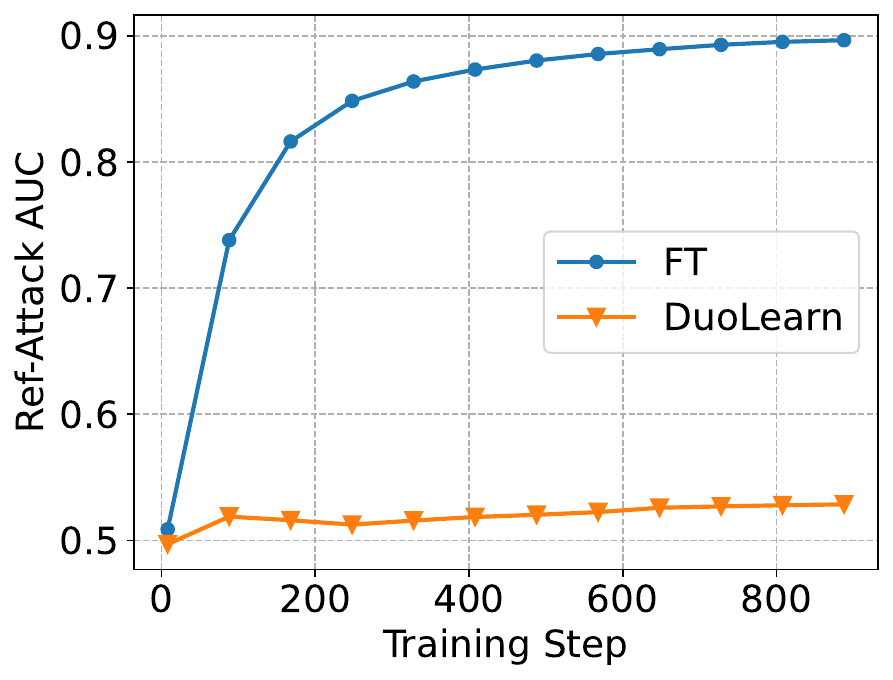}
    \includegraphics[width=0.33\linewidth]{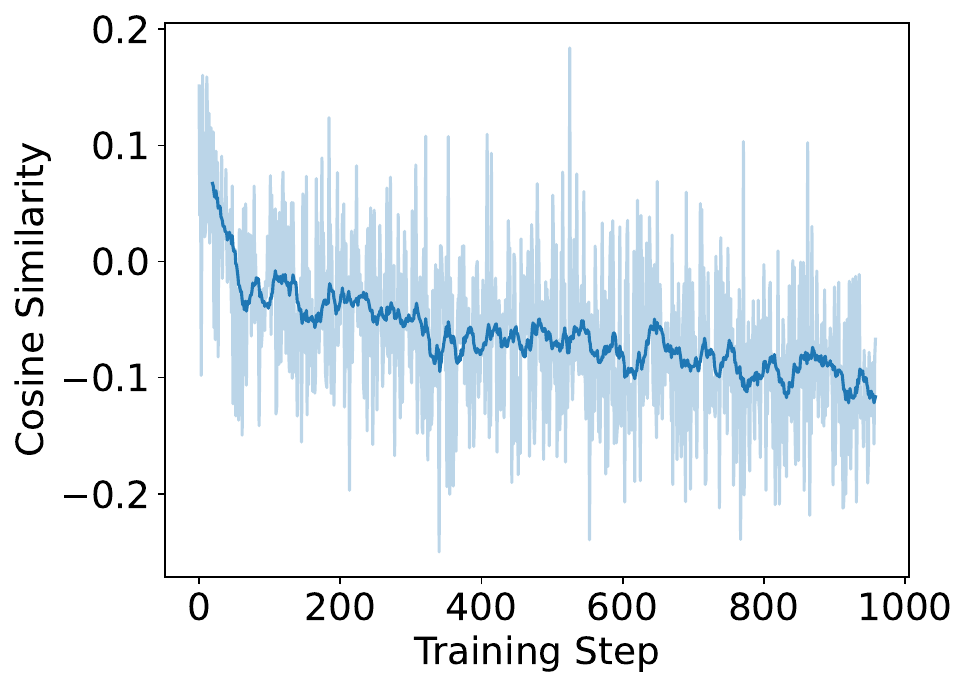}
    \caption{Training dynamics of \methodname and the conventional fine-tuning approach. The left and middle figures provide the training-testing gap and membership inference risks, respectively. The testing~$\mathcal{L}_{CE}$ of FT and training~$\mathcal{L}_{CE}$ of \methodname are significantly overlapping, we provide the breakdown in Figure~\ref{fig:add-overlap-breakdown} in Appendix~\ref{sec:app-add-res}. The right figure depicts the cosine similarity of the learning and unlearning gradients of \methodname. Cosine similarity of 1 means entire alignment, 0 indicates orthogonality, and -1 presents full conflict.}
    \label{fig:training-dynamics}
\end{figure*}

\textbf{Memorization and Generalization Dynamics}. Figure~\ref{fig:training-dynamics} (left) illustrates the training dynamics of conventional fine-tuning and \methodname, while Figure~\ref{fig:training-dynamics} (middle) depicts the membership inference risks. Generally, the gap between training and testing loss of conventional fine-tuning steadily increases over time, leading to model overfitting and high privacy risks. In contrast, \methodname maintains a stable equilibrium where the gap remains more than 10 times smaller. This equilibrium arises from the dual-purpose loss, which balances learning on hard tokens while actively unlearning memorized tokens. By preventing excessive memorization, \methodname mitigates membership inference risks and enhances generalization.

\noindent \textbf{Gradient Conflicts}. To study the conflict between the learning and unlearning objectives in our dual-purpose loss function, we compute the gradient for each objective separately. We then calculate the cosine similarity of these two gradients. Figure~\ref{fig:training-dynamics} (right) provides the cosine similarity between two gradients over time. During training, the cosine similarity typically ranges from -0.15 to 0.15. This indicates a mix of mild conflicts and near-orthogonal updates. On average, it decreases from 0.05 to -0.1. This trend reflects increasing gradient misalignment. Early in training, the model may not have strongly learned or memorized specific tokens, so the conflicts are weaker. Overtime, as the model learns more and memorization grows, the divergence between hard and memorized tokens increases, making the gradients less aligned. This gradient conflict is the root of the small degradation of language modeling performance of \methodname compared to the conventional fine-tuning approach.

\begin{figure}[htp]
    \centering
    \includegraphics[width=0.7\linewidth]{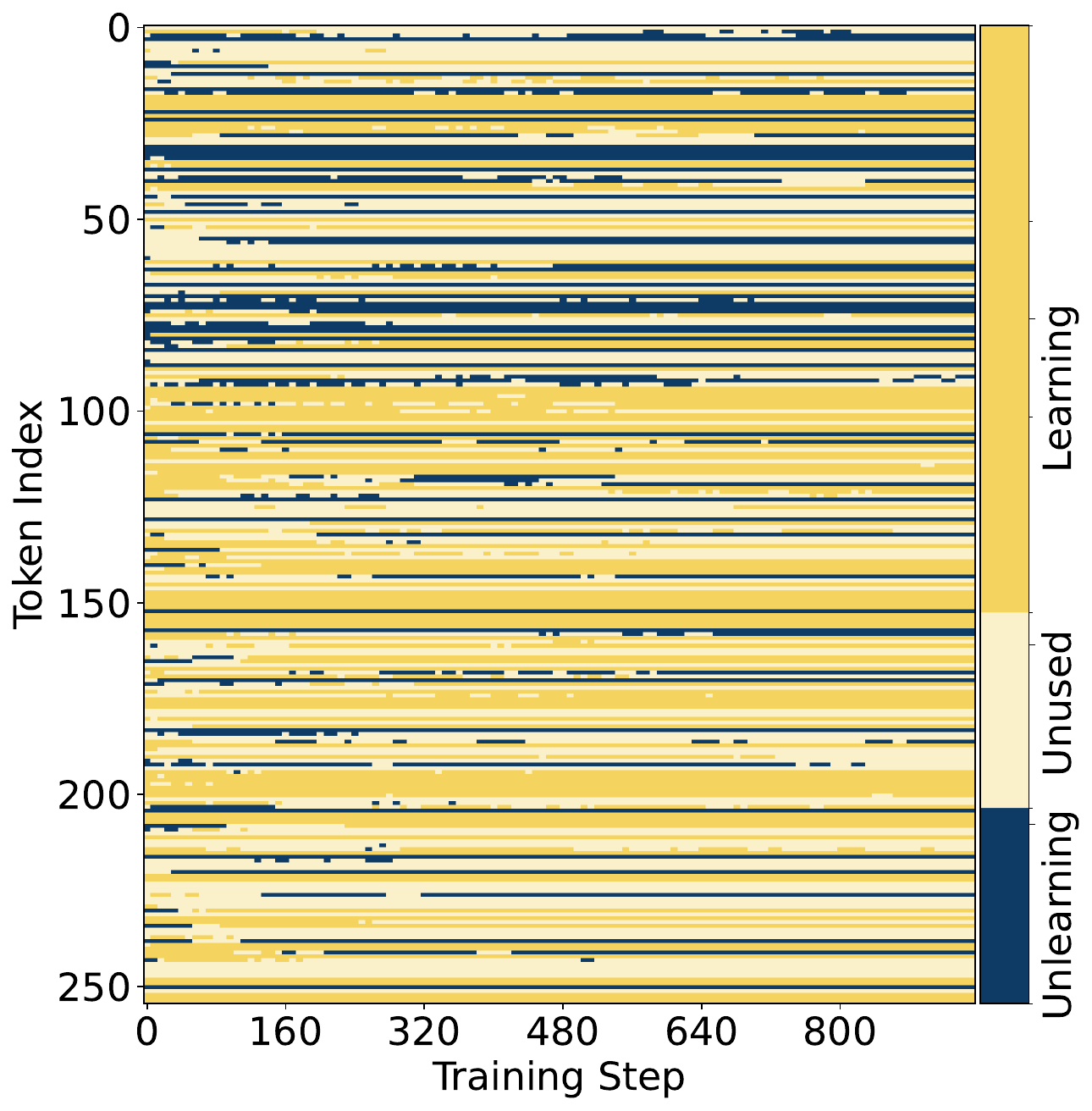}
    \caption{Token Selection Dynamics of \methodname}
    \label{fig:token-selection}
\end{figure}

\noindent \textbf{Token Selection Dynamics}. Figure~\ref{fig:token-selection} illustrates the token selection dynamics of \methodname during training. The figure shows that the token selection process is dynamic and changes over epochs. In particular, some tokens are selected for unlearning from the beginning to the end of the training. 
This indicates that a token, even without being selected as a learning token initially, can be learned and memorized through the connections with other tokens. This also explains that simple masking as in Goldfish is not sufficient to protect against MIAs. Additionally, there are a significant number of tokens that are selected for learning in the early epochs but selected for unlearning in the later epochs. This indicates that the model gradually memorizes these tokens over epochs, and the during-training unlearning process is essential to mitigate the memorization risks.

\subsection{Privacy Backdoor}
To study the worst case of privacy attacks and defense effectiveness under the state-of-the-art MIA, we perform a privacy backdoor -- Precurious~\cite{precurious}. In this setup, the target model undergoes continual fine-tuning from a warm-up model. The attacker then applies a reference-based MIA that leverages the warm-up model as the attack's reference. Table~\ref{tab:backdoor} shows the language modeling and MIA performance on CCNews with GPT-2. Precurious increases the MIA AUC score by 5\%. Goldfish achieves the lowest PPL, aligning with~\citet{hans2024be}, where the Goldfish masking mechanism acts as a regularizer that potentially enhances generalization. Both DPSGD and \methodname provide strong privacy protection, with \methodname offering slightly better defense while maintaining lower perplexity than DPSGD.


\begin{table}[h]
    \centering
    \resizebox{\columnwidth}{!}{\begin{tabular}{c|cccccc}
        \textbf{Metric} & \textbf{WU} & \textbf{FT} & \textbf{GF} & \textbf{DP} & \textbf{DuoL} \\ \hline
        \textbf{PPL} & \textit{23.318} & 21.593 & \textbf{21.074} & 23.279 & 22.296  \\
        \textbf{AUC} & \textit{0.500} & 0.911 & 0.886 & 0.533 & \textbf{0.499} \\
    \end{tabular}}
    \caption{Experimental results of privacy backdoor for GPT2 on the CC-news dataset. WU stands for the warm-up model leveraged by Precurious. GF, DP, and DuoL are abbreviations of Goldfish, DPSGD, and \methodname}
    \label{tab:backdoor}
\end{table}

\begin{table*}[!h]
    \centering
    \resizebox{0.85\textwidth}{!}{\begin{tabular}{|l|c|c|c|c|}
    \hline
    \textbf{Method} & \textbf{Eval Loss (↓)} & \textbf{MIA AUC (↓)} & \textbf{TPR @ 1\% FPR (↓)} & \textbf{Training Time (↓)} \\
    \hline
    \textit{Untrained model} & \textit{11.329} & \textit{0.517} & \textit{0.010} & \textit{N/A} \\
    Conventional LM & 3.003 & 0.930 & 0.376 & \textbf{10.095} hours \\
    Goldfish & \textbf{2.844} & 0.905 & 0.426 & 10.159 hours \\
    DuoLearn & 3.244 & \textbf{0.548} & \textbf{0.040} & 12.278 hours\\
    \hline
    \end{tabular}}
    \caption{Comparison of Methods on Evaluation Loss, Privacy Metrics, and Training Time for pretraining.}
    \label{tab:pretraining}
    \vspace{-3mm}    
\end{table*}

\subsection{Pretraining}
We conduct a small-scale pretraining experiment using a Llama-like architecture with 1.5 billion parameters. The experiment is to pretrain on a dataset of 1 billion tokens. We reuse the dataset and codebase developed by~\citet{sanyal2024inheritune}\footnote{Following this setting and due to limited computing resources, we use a batch size of 8 which is much smaller than practical pretraining. All methods implement the same learning rate and are evaluated at their 25K-th iteration.}. The pretraining corpus is collected from various sources and domains, including arXiv, books, Common Crawl, GitHub, StackExchange, and Wikipedia~\cite{weber2024redpajama}. To train the reference model, we use 10\% of the data. Table~\ref{tab:pretraining} presents the performance of DuoLearn in this pretraining. Generally, \methodname successfully mitigates the MIA risk, reducing AUC from 0.9 to 0.55 and TPR at 1\% FPR from 0.4 to 0.04, with minimal degradation on the model performance. 

\subsection{Low-Prevalence Tokens vs High-Prevalence Tokens}
\begin{figure}[h!]
    \centering
    \includegraphics[width=\linewidth]{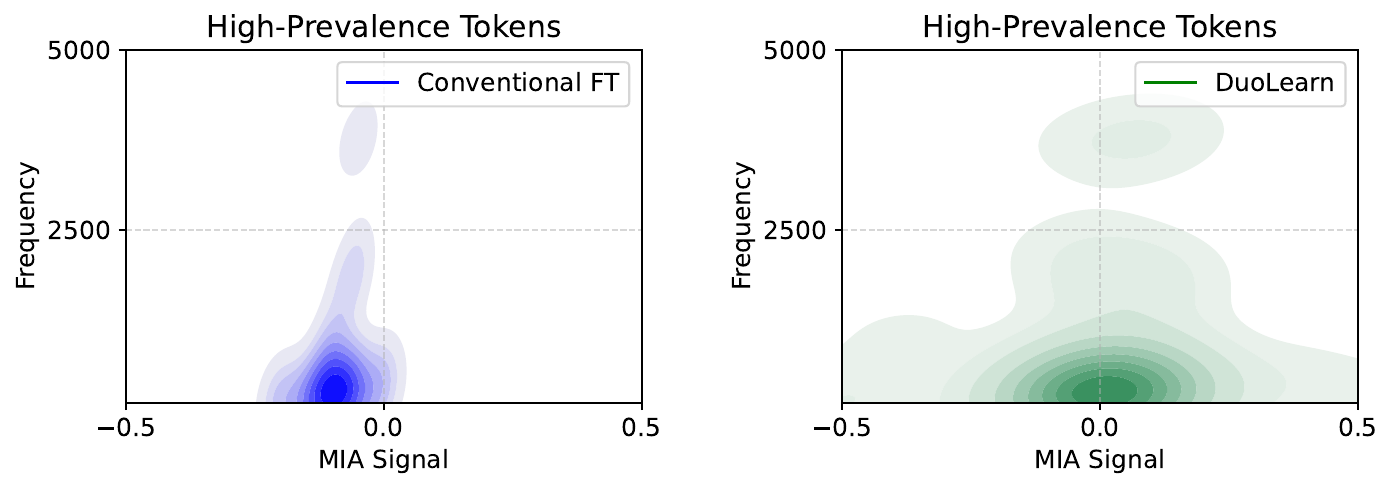} \\
    \includegraphics[width=\linewidth]{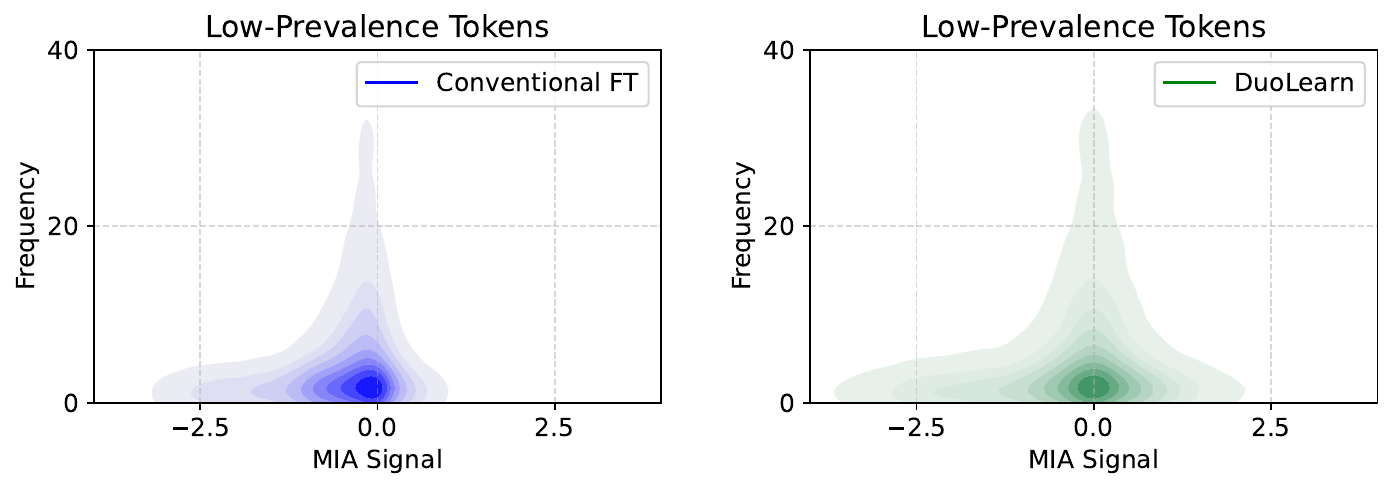} \\
    \caption{Effects on high-prevalence and low-prevalence tokens. For MIA signal, we consider the Ref-Loss attack, a signal value closer to zero indicates lower risk.}
    \label{fig:prevalance}
\end{figure}

To understand the bias of the methods towards low-prevalence and high-prevalence tokens, we conduct some visualization on the token-level MIA signals and token frequencies. We selected a set of 500 samples in the training data. 
We split the tokens into two sets, high-frequency and low-frequency, then visualize kernel density estimation (KDE) separately, illustrated in Figure~\ref{fig:prevalance}. Notably, both conventional FT and \methodname yield different privacy risks for the two groups. The low-prevalence tokens have higher per-token MIA risks (with MIA signal ranging from -5 to 2, compared to -0.5 to 0.5 of high-prevalence tokens). \methodname successfully shifts the MIA-signal distribution to be centered at zero for both high-frequency and low-frequency tokens. This indicates that \methodname reduces the majority of MIA risks across both token groups. However, it slightly increases the variance of risks for both groups.

\section{Conclusion}
We introduced \methodname, an effective training framework defending against MIAs for LLMs. The extensive experiments demonstrate its robustness in protecting privacy while maintaining strong language modeling performance across various datasets and architectures. Although our study focuses on fine-tuning and small scale pretraining due to computational constraints, \methodname can be seamlessly applied to large-scale pretraining, as in prior selective pretraining work~\cite{lin2024not}. By categorizing tokens and treating them appropriately, \methodname opens a novel pathway for MIA defense. Future work can explore improved token selection strategies and multi-objective training approaches.

\section*{Limitations}
The main limitation of our work is the small-scale experiment settings due to limited computing resources. However, we believe \methodname can be directly applied to large-scale pretraining without  any modifications, as  in previous reference-model-based pretraining study~\cite{lin2024not}. Another limitation is the reference model, which may be restrictive in highly sensitive or domain-limited settings~\cite{tramr2024position}. From a technical perspective, while \methodname performs well across different datasets and architectures, there is room for improvement. 
For example, future work could explore adaptive selection size or weighted token contribution. 
Additionally, as \methodname is an empirical defense,
future work can investigate the convergence and overfitting analysis.

\section*{Acknowledgment}
This work is partially supported by the National Science Foundation under Award Numbers 2302968, 2124104, and 2125530, and by the National Institutes of Health under Award Numbers R01ES033241 and R01LM013712. The views and opinions expressed in this paper are those of the authors and do not necessarily reflect the views of the U.S. Government or any agency thereof.

\bibliography{acl_latex}

\appendix

\section{Additional Related Works}
\label{sec:app-add-rel-works}
\subsection{Training Data Selection}

\begin{figure*}[!ht]
    \centering
    \includegraphics[width=\textwidth]{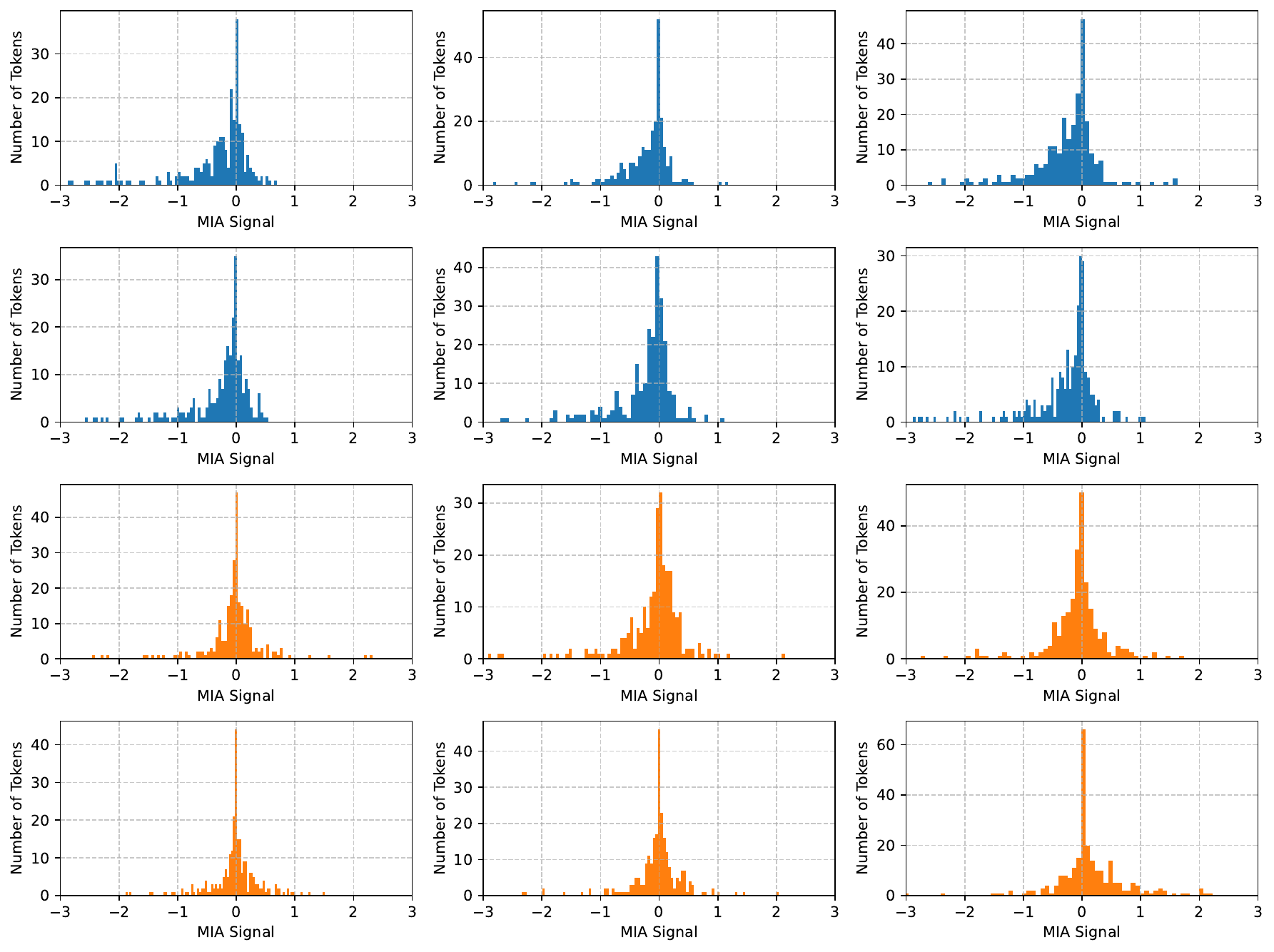}
    \caption{Histograms of MIA signal of tokens. Each figure depicts a sample. Blue means the member samples while orange represents the non-member samples. We limited the y-axis range to -3 to 3 for better visibility, so it can result in missing several non-significant outliers.}
    \label{fig:add-per-token-loss}
\end{figure*}

Training data selection are methods that filter high-quality data from noisy big data \textit{before training} to improve the model utility and training efficiency. There are several works leveraging reference models~\cite{Coleman2020Selection, xie2023doremi}, prompting LLMs~\cite{li-etal-2024-one}, deduplication~\cite{lee2022deduplicating, kandpal2022deduplicating}, and distribution matching~\cite{kang2024get}. However, we do not aim to cover this data selection approach, as it is orthogonal and can be combined with ours.

\subsection{Selective Training}
Selective training refers to methods that \textit{dynamically choose} specific samples or tokens \textit{during training}. Selective training methods are the most relevant to our work. Generally, sample selection has been widely studied in the context of traditional classification models via online batch selection~\cite{loshchilov2016o, Angelosonl, pmlr-v108-kawaguchi20a}. These batch selection methods replace the naive random mini-batch sampling with mechanisms that consider the importance of each sample mainly via their loss values. ~\citet{2022PrioritizedTraining} indeed choose highly important samples from regular random batches by utilizing a reference model. However, due to the sequential nature of LLMs, which makes the training significantly different from the traditional classification ML, sample-level selection is not effective for language modeling~\cite{kaddour2023no}. \citet{lin2024not} extend the reference model-based framework to select meaningful tokens within batches. All of the previous methods for selective training aim to improve the training performance and compute efficiency. Our work is the first looking at this aspect for defending against MIAs.

\section{Token-level membership inference risk analysis}
\begin{figure*}[!ht]
    \centering
    \includegraphics[width=0.28\textwidth]{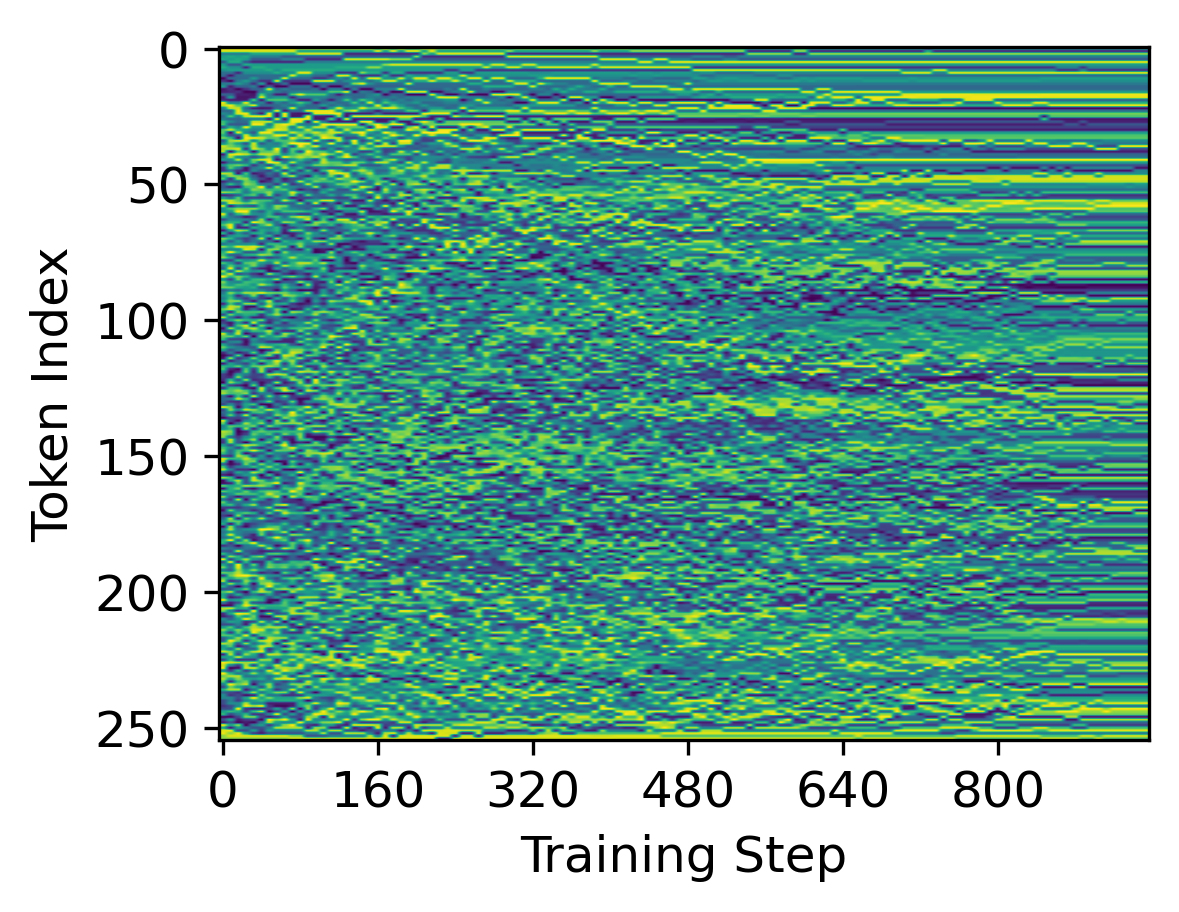}
    \includegraphics[width=0.28\textwidth]{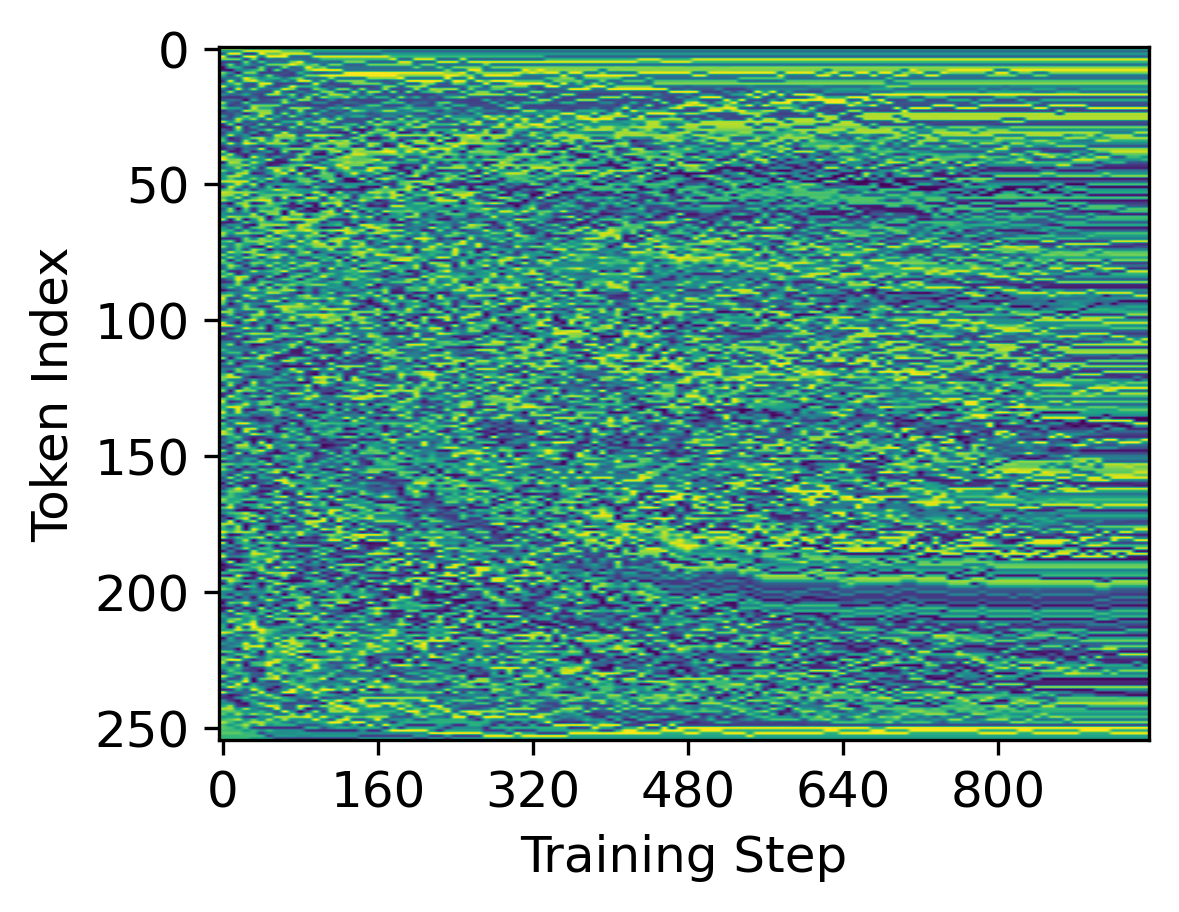}
    \includegraphics[width=0.3\textwidth]{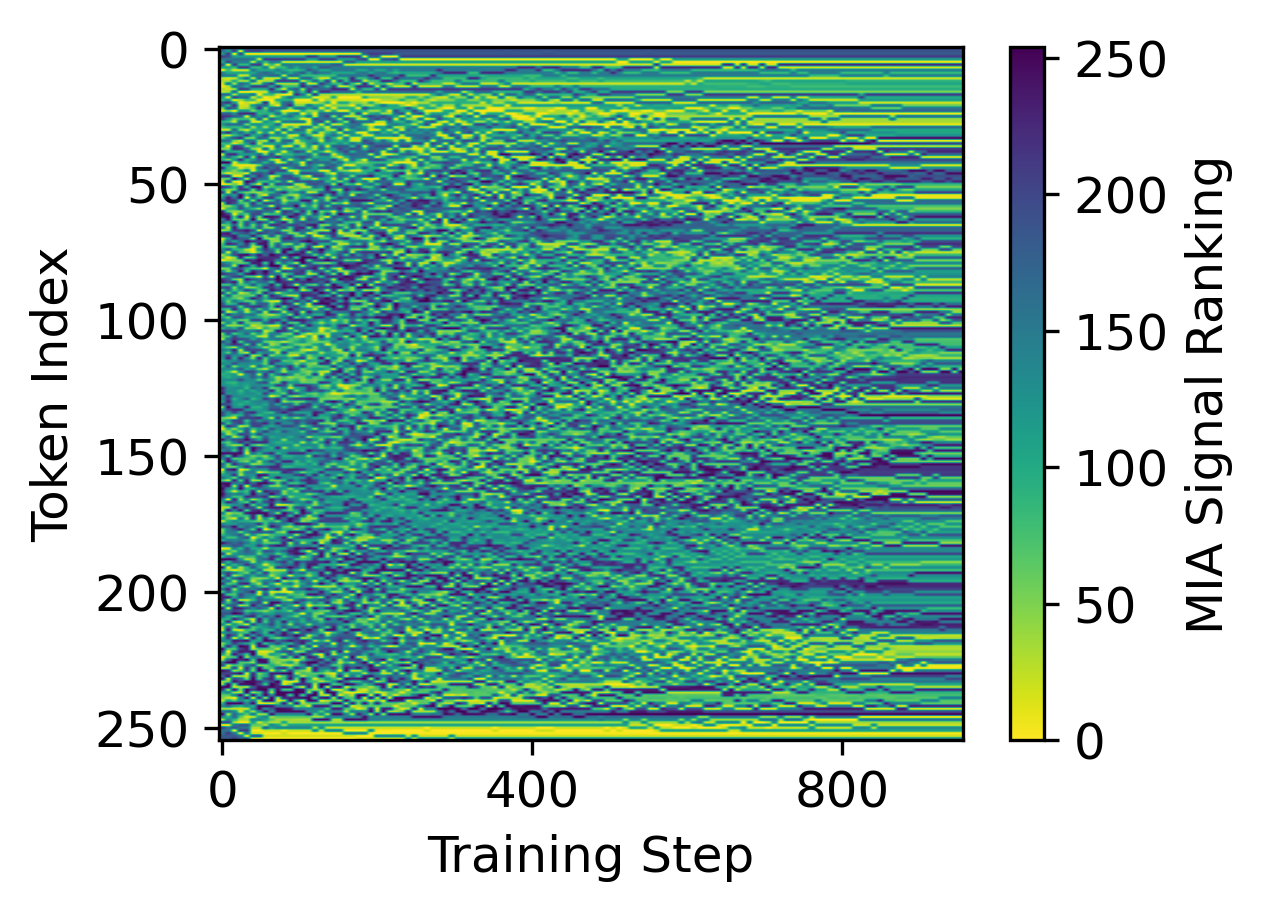}    
    \caption{MIA signal ranking of tokens during training. Each figure illustrates a sample.}
    \label{fig:add-per-token-dynamics}
\end{figure*}

Figures~\ref{fig:add-per-token-loss} and~\ref{fig:add-per-token-dynamics} present the analysis for additional samples. Generally, the trends are consistent with the one presented in Section~\ref{sec:analysis}.

\label{sec:app-analysis}

\section{Experiment settings}
\subsection{Implementation details}
\label{sec:app-implementation}
$\bullet$ \textbf{FT}. We implement the conventional fine tuning using Huggingface Trainer. We manually tune the learning rate to make sure no significant underfitting or overfitting. The batch size is selected appropriately to fit the physical memory and comparable with the other methods'.

\noindent $\bullet$ \textbf{Goldfish}. Goldfish is also implemented with Huggingface Trainer, where we custom the \texttt{compute\_loss} function. We implement the deterministic masking version rather than the random masking to make sure the same tokens are masked over epochs, potentially leading to better preventing memorization. The learning rate is also manually tuned, we noticed that the optimal Goldfish learning rate is usually slightly greater than FT's. This can be the gradients of two methods are almost similar, Goldfish just removes some tokens' contribution to the loss calculation. The batch size of FT can set as the same as FT, as Goldfish does not have significant overhead on memory.

\noindent $\bullet$ \textbf{DPSGD}. DPSGD is implemented by FastDP~\cite{bu2023zero}. We implement DPSGD with fastDP~\cite{bu2023zero} which offers state-of-the-art efficiency in terms of memory and training speed. We also use automatic clipping~\cite{bu2023automatic} and a mixed optimization strategy~\cite{mixclipping} between per-layer and per-sample clipping for robust performance and stability.

\noindent $\bullet$ \textbf{\methodname}. We implement \methodname using Huggingface Trainer, same as FT and Goldfish. The learning is reused from FT. The batch size of \methodname is usually smaller than FT and Goldfish when the model becomes large such as Pythia and Llama 2 due to the reference model, which consumes some memory.

For a fair comparison, we aim to implement the same batch size for all methods if feasible. In case of OOM (out of memory), we perform gradient accumulation, so all the methods can have comparable batch sizes. We provide the hyper-parameters of method for GPT2 in Table~\ref{tab:hyperparameter}. For Pythia and Llama 2, the learning rate, batch size, and number of epochs are tuned again, but the hyper-parameters regarding the privacy mechanisms remain the same. To make sure there is no naive overfitting, we evaluate the methods by selecting the best models on a validation set. Moreover, the testing and attack datasets remains identical for evaluating all methods. Additionally, we balance the number of member and non-member samples for MIA evaluation. It is worth noting that for the ablation study and analysis, if not state, the default model architecture and dataset are GPT2 and CC-news.

\begin{table*}[!ht]
    \centering
    \begin{tabular}{c|clc}
    \textbf{LLM} & \textbf{Method} & \textbf{Hyper-parameter} & \textbf{Value}  \\ \hline
     \multirow{22}{*}{\textbf{GPT2}}  &  \multirow{4}{*}{FT} &  Learning rate & 1.75e-5 \\ 
     & & Batch size & 96 \\
     & & Gradient accumulation steps & 1 \\
     & & Number of epochs & 20 \\ \cline{2-4}
       &  \multirow{5}{*}{Goldfish} &  Learning rate & 2e-5 \\ 
     & & Batch size & 96 \\
     & & Grad accumulation steps & 1 \\
     & & Number of epochs & 20 \\
     & & Masking Rate & 25\% \\ \cline{2-4}
           &  \multirow{6}{*}{DPSGD} &  Learning rate & 1.5e-3 \\ 
     & & Batch size & 96 \\
     & & Grad accumulation steps & 1 \\
     & & Number of epochs & 10 \\
     & & Clipping & automatic clipping \\ 
     & & Privacy budget & (8, 1e-5)-DP \\ \cline{2-4}
           &  \multirow{6}{*}{DuoLearn} &  Learning rate & 1.75e-3 \\ 
     & & Batch size & 96 \\
     & & Grad accumulation steps & 1 \\
     & & Number of epochs & 20 \\
     & & $K_h$ & 60\% \\ 
     & & $K_m$ & 20\% \\
     & & $\tau$ & 0 \\
     & & $\alpha$ & 0.8 \\ \hline
    \end{tabular}
    \caption{Hyper-parameters of the methods for GPT2.}
    \label{tab:hyperparameter}
\end{table*}

\section{Additional Results}
\label{sec:app-add-res}

\begin{figure}[h!]
    \centering
    \includegraphics[width=0.7\linewidth]{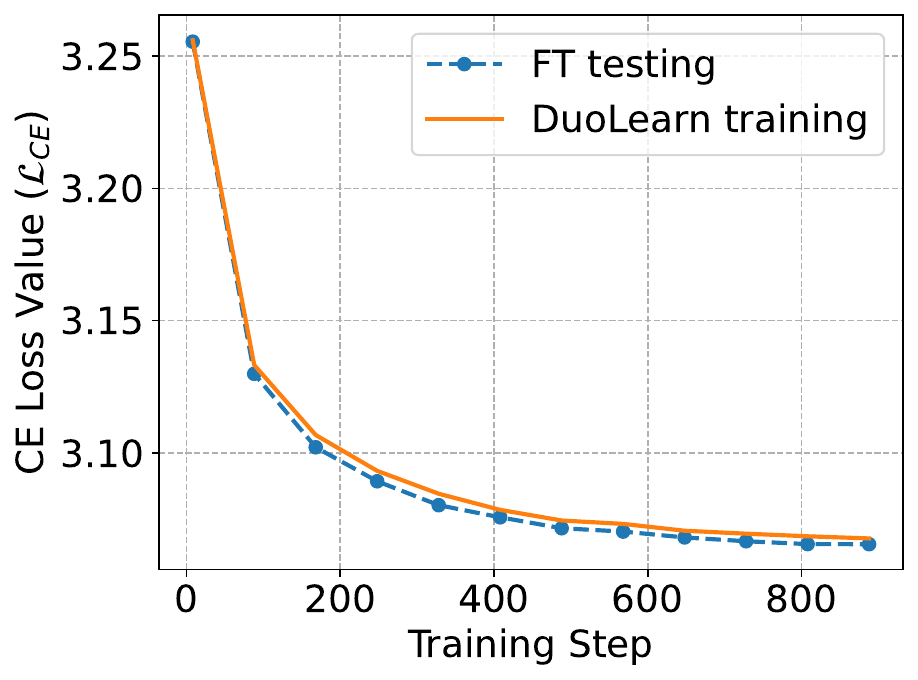}
    \caption{Breakdown to the cross entropy loss values of FT on the testing set and \methodname on the training set during training.}
    \label{fig:add-overlap-breakdown}
\end{figure}

\subsection{Overall Evaluation}

\begin{table*}[!ht]
  \centering
  \resizebox{\textwidth}{!}{\begin{tabular}{cl|ccccc|ccccc}
   \multirow{3}{*}{\textbf{LLM}}  & \multirow{3}{*}{\textbf{Method}} &  \multicolumn{5}{c|}{\textbf{Wikipedia}} & \multicolumn{5}{c}{\textbf{CC-news}} \\ \cmidrule(lr){3-7}  \cmidrule(lr){8-12}
    &  & PPL & Loss & Ref & min-k & \multicolumn{1}{c|}{zlib} & PPL & Loss & Ref & min-k & zlib \\ \midrule
    \multirow{4}{*}{GPT2} & \textit{Base} & \textit{34.429} & \textit{0.002} & \textit{0.014} & \textit{0.010} & \textit{0.002} & \textit{29.442} & \textit{0.018} & \textit{0.002} & \textit{0.022} & \textit{0.006} \\ 
    \multirow{4}{*}{124M} & FT & \textbf{12.729} & 0.018 & 0.574 & 0.016 & 0.014 & \textbf{21.861} & 0.030 & 0.026 & 0.016 & 0.016 \\
    & Goldfish & 12.853 & 0.018 & 0.632 & 0.016 & 0.010 & 21.902 & 0.030 & 0.024 & 0.028 & 0.016 \\
    & DPSGD & 18.523 & \textbf{0.004} & 0.036 & 0.018 & 0.006 & 26.022 & \textbf{0.018} & \textbf{0.004} & \textbf{0.018} & 0.008 \\
    & \methodname & 13.628 & 0.014 & \textbf{0.010} & \textbf{0.014} & \textbf{0.004} & 23.733 & 0.030 & 0.022 & 0.026 & \textbf{0.006} \\ \midrule
    
    \multirow{4}{*}{Pythia} & \textit{Base} & \textit{10.287} & \textit{0.002} & \textit{0.014} & \textit{0.006} & \textit{0.008} & \textit{13.973} & \textit{0.002} & \textit{0.008} & \textit{0.020} & \textit{0.014} \\ 
    \multirow{4}{*}{1.4B} & FT & \textbf{6.439} & 0.020 & 0.440 & 0.010 & 0.020 & 11.922 & 0.014 & 0.008 & 0.022 & 0.020 \\
    & Goldfish & 6.465 & 0.016 & 0.412 & 0.010 & 0.020 & \textbf{11.903} & 0.014 & 0.008 & 0.024 & 0.018 \\
    & DPSGD & 7.751 & \textbf{0.004} & \textbf{0.016} & {0.010} & \textbf{0.004} & 13.286 & \textbf{0.002} & \textbf{0.004} & \textbf{0.018} & \textbf{0.014} \\
    & \methodname & 6.553 & 0.008 & 0.030 & \textbf{0.006} & 0.006 & 12.670 & 0.004 & 0.020 & \textbf{0.018} & 0.016 \\ \midrule
    
    \multirow{4}{*}{Llama-2} & \textit{Base} & \textit{7.014} & \textit{0.006} & \textit{0.016} & \textit{0.016} & \textit{0.010} & \textit{9.364} & \textit{0.006} & \textit{0.006} & \textit{0.024} & \textit{0.006} \\ 
    \multirow{4}{*}{7B} & FT & \textbf{3.830} & 0.028 & 0.170 & 0.030 & 0.028 & \textbf{6.261} & 0.002 & 0.018 & 0.002 & 0.002 \\
    & Goldfish & 3.839 & 0.028 & 0.198 & 0.028 & 0.028 & 6.280 & 0.002 & 0.018 & 0.002 & 0.006 \\
    & DPSGD & 4.490 & \textbf{0.006} & 0.014 & \textbf{0.020} & \textbf{0.010} & 6.777 & 0.008 & 0.026 & 0.016 & 0.010 \\
    & \methodname & 4.006 & 0.010 & \textbf{0.002} & 0.028 & 0.012 & 6.395 & \textbf{0.002} & \textbf{0.020} & \textbf{0.004} & \textbf{0.002} \\ 
  \end{tabular}}
  \caption{Overall Evaluation: Perplexity (PPL) and TPR at FPR of 1\% scores of the MIAs with different signals (Loss/Ref/Min-k/Zlib). For all metrics, the lower the value, the better the result.}
  \label{tab:tpr}
\end{table*}
Table~\ref{tab:tpr} provides the True Positive Rate (TPR) at low False Positive Rate (FPR) of the overall evaluation. Generally, compared to CC-news, Wikipedia poses a significant higher risk at low FPR. For example, the reference-based attack can achieve a score of 0.57~ on GPT2 if no protection. In general, Goldfish fails to mitigate the risk in this scenario, while both DPSGD and \methodname offer robust protection.

\subsection{Auxiliary dataset}
\begin{figure}[h!]
    \centering
    \includegraphics[width=0.8\linewidth]{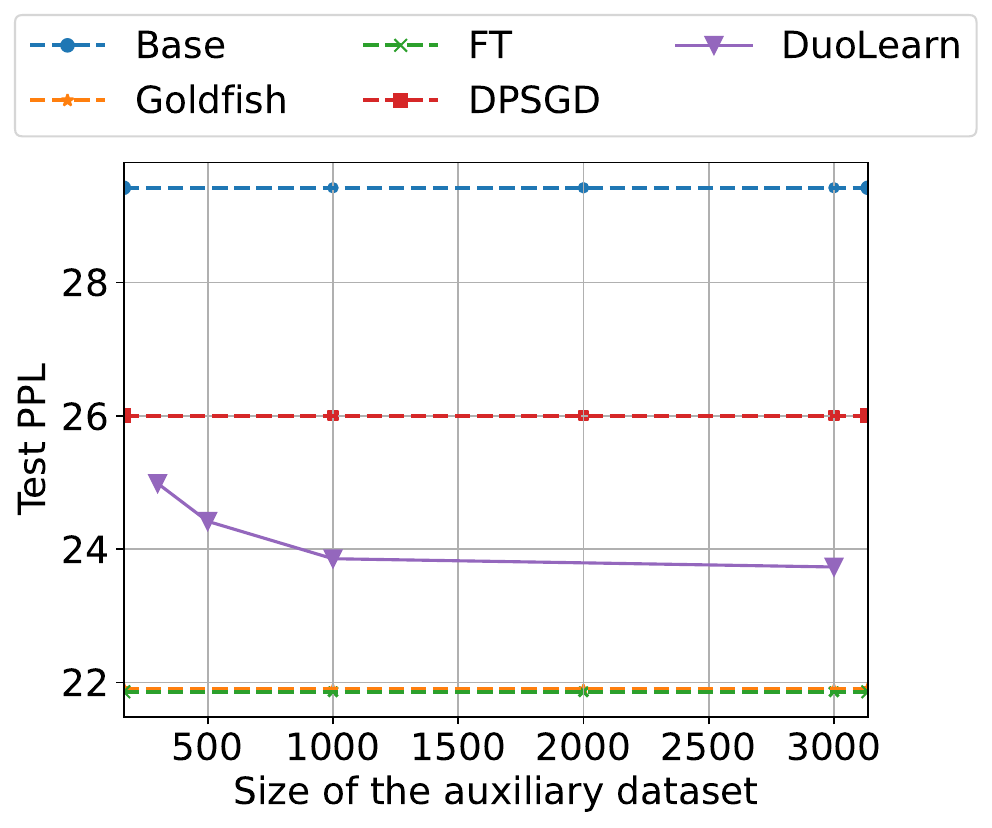}
    \caption{Language modeling performance while varying the auxiliary dataset's size. Note that the results of FT and Goldfish are significantly overlapping.}
    \label{fig:aux_size}
\end{figure}

We investigate the size of the auxiliary dataset which is disjoint with the training data of the target model and the attack model. In this experiment, the other methods are trained with 3K samples. Figure~\ref{fig:aux_size} presents the language modeling performance while varying the auxiliary dataset's size. The result demonstrates that the better reference model, the better language modeling performance. It is worth noting that even with a very small number of samples, \methodname can still outperform DPSGD. Additionally, there is only a little benefit when increasing from 1000 to 3000, this indicates that the reference model is not needed to be perfect, as it just serves as a calibration factor. This phenomena is consistent with previous selective training works~\cite{lin2024not, 2022PrioritizedTraining}.

\subsection{Hyperparameter sensitivity analysis}
\begin{figure}[h!]
    \centering
    \includegraphics[width=0.8\linewidth]{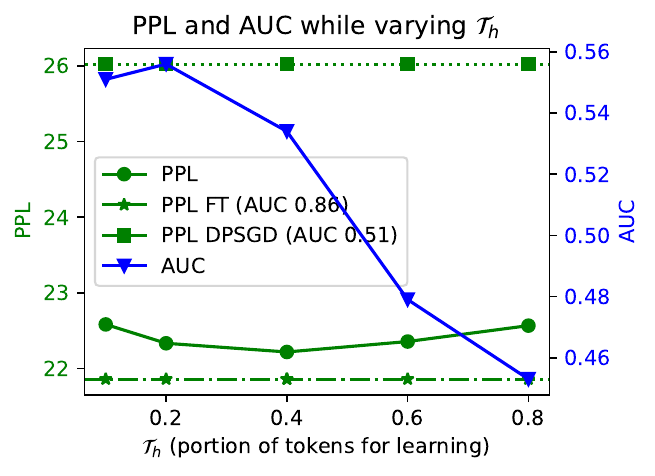}
    \caption{Performance of \methodname while varying $\mathcal{T}_{h}$}
    \label{fig:varying-tauh}
\end{figure}

\begin{figure}[h!]
    \centering
    \includegraphics[width=0.8\linewidth]{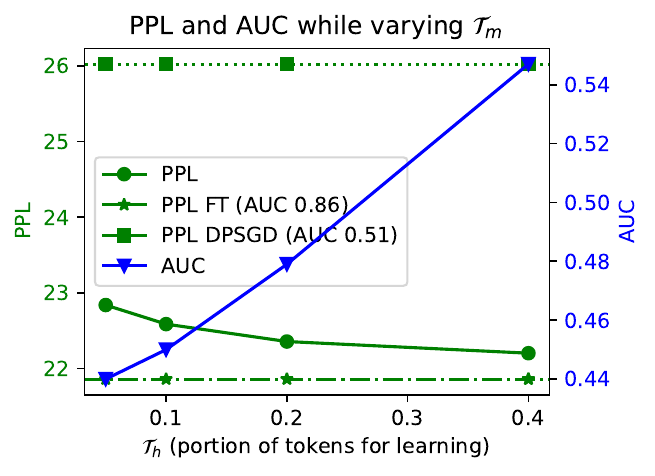}
    \caption{Performance of \methodname while varying $\mathcal{T}_{m}$}
    \label{fig:varying-taum}
\end{figure}

\textbf{$\bullet$ Varying $\mathcal{T}_{h}$ and $\mathcal{T}_{m}$} -- Portion of tokens for learning and for unlearning. We keep other hyperparamters as default and adjust $\mathcal{T}_{h}$ and $\mathcal{T}_{m}$ separately. Figures~\ref{fig:varying-tauh} and~\ref{fig:varying-taum} provides the results of this experiment. Generally, \methodname is robust while varying these hyperparameters with PPL ranging from 22 to 23 and AUC ranging from 0.46 to 0.56.

\begin{figure}[h!]
    \centering
    \includegraphics[width=0.8\linewidth]{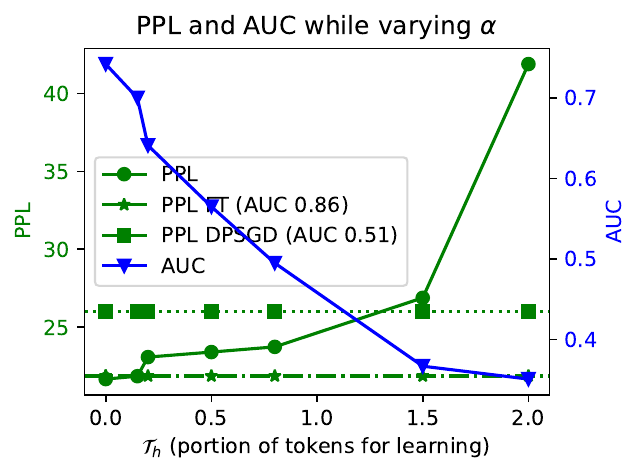}
    \caption{Performance of \methodname while varying $\alpha$}
    \label{fig:varying-alpha}
\end{figure}

\textbf{$\bullet$ Varying $\alpha$} -- Weight balance factor. 
Figure~\ref{fig:varying-alpha} illustrates the peformance while varying $\alpha$. Intuitively, the smaller $\alpha$, the less unlearining performed, it leads to a better language modeling performance and higher privacy risk. When $\alpha$ is unreasonably high (i.e., 1.5 or 2.0), the unlearning part dominates the learning one, it leads to high perplexity values of language modeling.  

\subsection{Training time}
We report the training time for full fine-tuning Pythia 1.4B. We manually increase the batch size that could fit into the GPU's physical memory. As a results, FT and Goldfish can run with a batch size of 48, while DPSGD and \methodname can reach the batch size of 32. We also implement gradient accumulation, so all the methods can have the same virtual batch size.

\begin{table}[!ht]
    \centering
    \begin{tabular}{c|c}
        \textbf{Training Time} & \textbf{\textbf{1 epoch}} (in minutes) \\ \hline
        {FT} & 2.10 \\ 
        {Goldfish} & 2.10 \\
        {DPSGD} & 3.19 \\ 
        {DuoLearn} & 2.85 
    \end{tabular}
    \caption{Training time for one epoch of (full) Pythia 1.4B on a single H100 GPU}
    \label{tab:training-time}
\end{table}

Table~\ref{tab:training-time} presents the training time for one epoch. Goldfish has little to zero overhead compared to FT. DPSGD and \methodname have a slightly higher training time due to the additional computation of the privacy mechanism. In particular, DPSGD has the highest overhead due to the clipping and noise addition mechanisms. Meanwhile, \methodname requires an additional forward pass on the reference model to select the learning and unlearning tokens. \methodname is also feasible to work at scale that has been demonstrated in the pretraining settings of the previous work~\cite{lin2024not}.

\end{document}